\documentclass[10pt,twocolumn,letterpaper]{article}

\usepackage{cvpr}              
\usepackage[accsupp]{axessibility} 
\usepackage{graphicx}
\usepackage{amsmath}
\usepackage{amssymb}
\usepackage{booktabs}

\usepackage[pagebackref,breaklinks,colorlinks]{hyperref}

\makeatletter
\DeclareRobustCommand\onedot{\futurelet\@let@token\@onedot}
\def\@onedot{\ifx\@let@token.\else.\null\fi\xspace}

\def\eg{\emph{e.g}\onedot} \def\Eg{\emph{E.g}\onedot}
\def\ie{\emph{i.e}\onedot} 
 
\def\etc{\emph{etc}\onedot} 
 
\def\etal{\emph{et al}\onedot}

\makeatother

\usepackage{xspace}
\newcommand{\name}[0]{SVRE\xspace}
\usepackage{color}

\usepackage[capitalize]{cleveref}
\crefname{section}{Sec.}{Secs.}
\Crefname{section}{Section}{Sections}
\Crefname{table}{Table}{Tables}
\crefname{table}{Tab.}{Tabs.}

\def\cvprPaperID{10026} 
\def\confName{CVPR}
\def\confYear{2022}

\usepackage{bm}
\usepackage{algorithm}
\usepackage{amsmath}
\usepackage{algorithmic}
\usepackage{multirow} 
\usepackage{xcolor}

\usepackage{overpic}
\usepackage{enumitem} 
\usepackage{overpic} 
\usepackage{color}

\definecolor{turquoise}{cmyk}{0.65,0,0.1,0.3}
\definecolor{purple}{rgb}{0.65,0,0.65}
\definecolor{dark_green}{rgb}{0, 0.5, 0}
\definecolor{orange}{rgb}{0.8, 0.6, 0.2}
\definecolor{red}{rgb}{0.8, 0.2, 0.2}
\definecolor{darkred}{rgb}{0.6, 0.1, 0.05}
\definecolor{blueish}{rgb}{0.0, 0.3, .6}
\definecolor{light_gray}{rgb}{0.7, 0.7, .7}
\definecolor{pink}{rgb}{1, 0, 1}
\definecolor{greyblue}{rgb}{0.25, 0.25, 1}

\usepackage{blindtext}

\renewcommand{\paragraph}[1]{\vspace{1em}\noindent\textbf{#1}.}

\usepackage{xspace}

\makeatletter
\DeclareRobustCommand\onedot{\futurelet\@let@token\@onedot}
\DeclareRobustCommand\onedot{\futurelet\@let@token\@onedot}
\def\@onedot{\ifx\@let@token.\else.\null\fi\xspace}

\def\eg{\emph{e.g}\onedot} \def\Eg{\emph{E.g}\onedot}
\def\ie{\emph{i.e}\onedot} 
 
\def\etc{\emph{etc}\onedot} 
 
\def\etal{\emph{et al}\onedot}

\begin{document}
\title{Stochastic Variance Reduced Ensemble Adversarial Attack 
\\for Boosting the Adversarial Transferability}


\author{Yifeng Xiong$^{1}$\thanks{The first two authors contribute equally.}, Jiadong Lin$^{1}$,  Min Zhang$^{1}$, John E. Hopcroft$^{2}$, Kun He$^{1}$\thanks{Corresponding author.}\\
$^{1}$Department of Computer Science, Huazhong University of Science and Technology, Wuhan, China \\
$^{2}$Department of Computer Science, Cornell University, Ithaca, NY, USA \\
{\tt\small\{xiongyf,jdlin,m\_zhang\}@hust.edu.cn},  {\tt\small jeh@cs.cornell.edu}, 
{\tt\small brooklet60@hust.edu.cn}
}


\maketitle

\begin{abstract}
The black-box adversarial attack has attracted impressive attention for its practical use in the field of deep learning security. Meanwhile, it is very challenging as there is no access to the network architecture or internal weights of the target model. Based on the hypothesis that if an example remains adversarial for multiple models, then it is more likely to transfer the attack capability to other models, the ensemble-based adversarial attack methods are efficient and widely used for black-box attacks. However, ways of ensemble attack are rather less investigated, and existing ensemble attacks simply fuse the outputs of all the models evenly. In this work, we treat the iterative ensemble attack as a stochastic gradient descent optimization process, in which the variance of the gradients on different models may lead to poor local optima. 
To this end, we propose a novel attack method called the stochastic variance reduced ensemble (SVRE) attack,  
which could reduce the gradient variance of the ensemble models and take full advantage of the ensemble attack. 
Empirical results on the standard ImageNet dataset demonstrate that the proposed method could boost the adversarial transferability and outperforms existing ensemble attacks significantly.
Code is available at \url{https://github.com/JHL-HUST/SVRE}.
\end{abstract}

\section{Introduction}
\label{section:introduction}

Deep neural networks (DNNs) have shown impressive performance on various computer vision tasks. However, recent researches have shown that DNNs are strikingly vulnerable to adversarial examples crafted by adding human-imperceptible perturbations~\cite{Szegedy14Intriguing, Goodfellow2014Explaining,papernot2016limitations}. 
Moreover, adversarial examples are known to be transferable that the examples crafted for one model can also mislead other black-box models~\cite{papernot2017practical,liu2017delving,moosavi2017universal}. 
Generating adversarial examples, \ie, adversarial attack, has drawn enormous attention since it can help evaluate the robustness of different models~\cite{carlini2017towards, tramer2020adaptive} and improve their robustness by adversarial training~\cite{Goodfellow2014Explaining,madry2018towards}.

Various adversarial attack methods have been proposed, including the optimization-based methods such as box-constrained L-BFGS~\cite{Szegedy14Intriguing} and Carlini \& Wagner’s method~\cite{carlini2017towards}, the gradient-based methods such as Fast Gradient Sign Method~\cite{Goodfellow2014Explaining} and its iterative variants~\cite{Kurakin2017physical,madry2018towards}, \etc.
In general, these adversarial attack methods can achieve high attack success rates in the white-box setting~\cite{carlini2017towards}, where the attacker can access the complete information of the target model, including the model architecture and gradient information.
However, these methods often exhibit low attack success rates in the black-box setting~\cite{dong2018boosting}, where the attacker can not access the information of the target model. In this case, the attacker either utilizes the transferability of adversarial examples to fool the black-box model, or attacks directly based on a small amount of queries on the output of the black-box model.

In recent years, a number of methods have been proposed to enhance the transferability of adversarial examples so as to improve the attack success rates in the black-box setting, including the gradient optimization attacks~\cite{dong2018boosting, lin2020nesterov, VT}, input transformation attacks~\cite{dong2019evading, xie2019improving, lin2020nesterov}, and model ensemble attacks~\cite{liu2017delving, dong2018boosting}. 
Among these methods, the model ensemble attacks are efficient and have been broadly adopted in boosting the black-box attack performance~\cite{xie2019improving, lin2020nesterov, gao2020patch}.
However, as compared to the other two categories that have been explored in depth, the category of model ensemble attack is rather less investigated. 

In this work, we observe that the existing model ensemble attack methods simply fuse the outputs of all models directly but ignore the variance of the gradients on different models, which may limit the potential capability of the model ensemble attacks.
Due to the inherent difference of the model architectures, the optimization paths of the models may differ widely, indicating that there exists considerable difference on the variance of the gradient directions among the possible models. 
Such variance may cause the optimization direction of the ensemble attack to be less accurate. 
As a result, the attack capability of the transferred adversarial examples is rather limited. 

To address this issue, we propose a novel method called the stochastic variance reduced ensemble (\name) attack 
to enhance the adversarial transferability of ensemble attacks. 
Our method is inspired by the stochastic variance reduced gradient (SVRG) method~\cite{johnson2013accelerating} designed for stochastic optimization, which has an outer loop that maintains an average gradient on a batch of data and an inner loop that randomly draws an instance from the batch and calculates an unbiased estimate of gradient based on the variance reduction.
In our method, we regard the ensemble models as the batch of data for the outer loop and randomly draw a model at each iteration of the inner loop. 
Taking the benign image as the initial adversarial example, 
the outer loop calculates the average gradient on the batch of models, and copies the current example to the inner loop, 
then the inner loop conducts multiple iterations of inner adversarial example updates. 
At each inner iteration, \name calculates the current gradient on a randomly picked model, tuned by 
the gradient bias of the outer adversarial example on this randomly picked model and on the ensemble model.  
At the end of the inner loop, the outer adversarial example is updated using the tuned gradient of the newest inner adversarial example. 

In this way, \name can obtain a more accurate gradient update at the outer loop to escape from poor local optima such that the crafted adversarial example would not ``overfit'' the ensemble model. 
Hence, the crafted adversarial example is expected to have higher transferability to other unknown models. 
To our knowledge, this is the first work to investigate the limitation of existing ensemble attack through the lens of gradient variance on multiple models. 
Extensive experiments on the ImageNet dataset demonstrate that \name consistently outperforms the vanilla ensemble model attack in the black-box setting.

\section{Related Works}
\label{section:related-works}

Let $x$ and $y$ be a benign image and the corresponding true label, respectively. Let $J(x,y)$ be the loss function of the classifier and $\mathcal{B}_\epsilon(x) = \{x':\|x-x'\|_p \leq \epsilon\}$ be the $L_p$-norm ball centered at $x$ with radius $\epsilon$. The goal of non-targeted adversarial attacks is to search for an adversarial example $x^{adv} \in \mathcal{B}_\epsilon(x)$ that maximizes the loss $J(x^{adv},y)$. To align with previous works, we focus on $L_\infty$-norm non-targeted adversarial attacks.

\subsection{Adversarial Attacks}
Existing adversarial attack 
methods can be categorized into three groups, namely gradient optimization attacks~\cite{Goodfellow2014Explaining,Kurakin2017physical, dong2018boosting,lin2020nesterov,VT}, input transformation attacks~\cite{dong2018boosting,xie2019improving,lin2020nesterov,DBLP:journals/corr/abs-2102-00436}, and model ensemble attacks~\cite{liu2017delving, dong2018boosting}.
\par

\textbf{Gradient optimization attacks.} ~The most typical adversarial attack based on gradient is the Fast Gradient Sign Method (FGSM)~\cite{Goodfellow2014Explaining}, which uses the gradient direction of the loss function with respect to the input image to generate a fixed amount of perturbation. Kurakin \etal \cite{Kurakin2017physical} propose the Basic Iterative Method (BIM) to run multiple iterations of FGSM with a small perturbation. Madry \etal \cite{madry2018towards} propose a noisy version of BIM, named the Projected Gradient Descent (PGD). Although PGD exhibits good attack performance in the white-box setting~\cite{athalye2018obfuscated}, it overfits the target model easily and yields weak transferability in the black-box setting. In order to improve the transferability of adversarial attacks, Dong \etal \cite{dong2018boosting} propose to boost the adversarial attack with momentum. More recently, Lin \etal \cite{lin2020nesterov} introduce Nesterov accelerated gradient method into the gradient-based attack to look ahead effectively to avoid overfitting. Wang \etal \cite{VT} reduce the variance of the gradient at each iteration to stabilize the update direction.
\par
\textbf{Input transformation attacks.} Another line of attacks focuses on adopting various input transformations to further improve the transferability of adversarial examples. Xie \etal \cite{xie2019improving} propose the Diverse Input Method (DIM) \cite{xie2019improving}, which utilizes random resizing and padding to create diverse input patterns to generate adversarial examples.  Dong \etal \cite{dong2019evading} propose the Translation-Invariant Method (TIM), 
which optimizes the perturbation over a set of translated images. Lin \etal \cite{lin2020nesterov} discover the scale-invariant property of deep learning models and propose the Scale-Invariant Method (SIM), which optimizes the adversarial perturbations over the scale copies of the input images.  Wang \etal ~\cite{DBLP:journals/corr/abs-2102-00436} propose the Admix, that calculates the gradient on the
input image admixed with a small portion of each add-in
image to craft more transferable adversaries.
\par
\textbf{Model ensemble attacks.} Liu \etal ~\cite{liu2017delving} find that attacking multiple models simultaneously can also improve the attack transferability. They fuse the predictions of multiple models to get the loss of ensemble predictions and adopt existing adversarial attacks (\eg FGSM and PGD) to generate adversarial examples using the loss. Dong \etal \cite{dong2018boosting} propose two variants of the model ensemble attack, namely fusing the logits and fusing the losses, respectively. Compared with various explorations on gradient optimization or input transformation, the model ensemble attacks are far less investigated, and existing methods only simply fuse the output predictions, logits, or losses.

\subsection{Adversarial Defenses}
As the counterpart of adversarial attacks, various 
defense methods have also been proposed, including  
adversarial training based defenses~\cite{Szegedy14Intriguing,ensmodel, madry2018towards, song2018improving,xie2019feature, ZhaiAdversarially,Song2020Robust,Gong_2021_CVPR} and input transformation based defenses~\cite{defense-JPEG,defense-RP,defense-HGD,defense-Bit-red,defense-ComDefend,defense-FD,defense-RS,defense-NRP}.

\textbf{Adversarial training based defenses.} Adversarial training is considered one of the most efficacious defense approaches which augments the training data by generating adversarial examples during the training process. Tram{\`{e}}r \etal~\cite{ensmodel} propose ensemble adversarial training, which augments the training data with perturbations transferred from other models. Madry \etal ~\cite{madry2018towards} propose PGD-Adversarial Training (PGD-AT), which augments the training data with adversarial examples crafted by PGD attack. Xie \etal~\cite{xie2019feature} develop new network architectures that increase adversarial robustness by performing feature denoising.
Adversarial training, while promising, is computationally expensive and hard to scale to large-scale datasets~\cite{Kurakin2017scale}.

\textbf{Input transformation based defenses.} This line of defenses aims to diminish the adversarial perturbations from the input data. Guo \etal~\cite{defense-JPEG} and Xie \etal~\cite{defense-RP} conduct transformations on images to remove the adversarial perturbations. Liao \etal~\cite{defense-HGD} use high-level representation guided denoiser (HGD) to purify the adversarial images. Xu \etal~\cite{defense-Bit-red} propose two feature squeezing methods, \ie bit reduction (Bit-R) and spatial smoothing to detect adversarial examples. Liu \etal~\cite{defense-FD} propose the feature distillation (FD), which adopts a JPEG-based defensive compression framework to diminish the adversarial perturbations. Jia \etal~\cite{defense-ComDefend} utilizes an end-to-end image compression model named ComDefend to defend against adversarial examples. Jia \etal~\cite{defense-RS} leverage the randomized smoothing (RS) to train a certifiably robust ImageNet classifier. Naseer \etal~\cite{defense-NRP} develop a neural representation purifier (NRP) model, which learns to purify the adversarially perturbed images through automatically derived supervision.

\section{Methodology}
\label{section:methodology}
We focus on addressing the adversarial transferability through the lens of reducing the gradient variance of the ensemble models used for crafting the adversarial example.  
Since our method is based on the model ensemble attack, we first introduce the existing ensemble attack methods, then present our motivation and elaborate the proposed \name in detail.

\subsection{Ensemble Attack Methods}
\label{section:methodology-1}
The ensemble attack\cite{liu2017delving,dong2018boosting} is an effective strategy to enhance the adversarial transferability. 
Its basic idea is to generate the adversarial examples using multiple models.

\textbf{Ensemble on predictions.}~ Liu \etal \cite{liu2017delving} first propose to achieve an ensemble attack by averaging the predictions (predicted probability) of the models. For an ensemble of $K$ models, the loss function of the ensemble model is:
   \begin{equation}
\label{eq:ens_on_predictions}
    J(\bm{x}, y) = -\mathbf{1}_{y}\cdot\log(\textstyle\sum_{k=1}^K w_k \bm{p}_k(\bm{x})),
\end{equation}
where $\mathbf{1}_{y}$ is the one-hot encoding of the ground-truth label $y$ of $\bm{x}$, $\bm{p}_k$ is the prediction of the $k$-th model, and $w_k \geq 0$ is the ensemble weight constrained by $\textstyle\sum_{k=1}^K w_k = 1$. 

\textbf{Ensemble on logits.}~ Dong \etal \cite{dong2018boosting} propose to fuse the logits (output before softmax) of models. For the ensemble of $K$ models, the loss function ensembled on logits is: 
\begin{equation}
\label{eq:ens_on_logits}
    J(\bm{x}, y) = -\mathbf{1}_{y}\cdot\log(\mathrm{softmax}( \textstyle\sum_{k=1}^K w_k \bm{l}_k(\bm{x}))),
\end{equation}
where $\bm{l}_k$ is the logits of the $k$-th model.

\textbf{Ensemble on losses.}~ Dong \etal \cite{dong2018boosting} also introduce an alternative ensemble attack 
by averaging the loss of $K$ models as follows:  
\begin{equation}
\label{eq:ens_on_loss}
    J(\bm{x}, y) = \textstyle\sum_{k=1}^K w_k J_k(\bm{x}, y),
\end{equation}
where $J_k$ is the loss of the $k$-th model.

For the weight parameters, the three methods simply choose the average weight in experiments, \ie $w_k = 1/K$.

\subsection{Rethinking the Ensemble Attack}
\label{section:methodology-2}
The ensemble attack method has been broadly adopted in enhancing the performance of black-box attacks~\cite{liu2017delving,dong2018boosting,xie2019improving, lin2020nesterov, gao2020patch,VT}. 
However, to our knowledge, researchers only utilize the existing ensemble attack strategy as a plug-and-play module to enhance their own attack methods, but did not delve into the ensemble attack method itself.

Intuitively, existing ensemble attack methods~\cite{dong2018boosting,liu2017delving} are helpful in improving the adversarial transferability because attacking an ensemble model can help to find better local maxima and makes it easier to generalize to other black-box models. However, merely averaging the outputs (logits, predictions or loss) of the models to build an ensemble model for the adversarial attack may limit the attack performance, 
as the individual optimization path of different model may vary diversely, but the variance is not considered, leading to an overfit to the ensemble model. 

\begin{figure}[tb]
\centering
\includegraphics[width=0.8\linewidth]{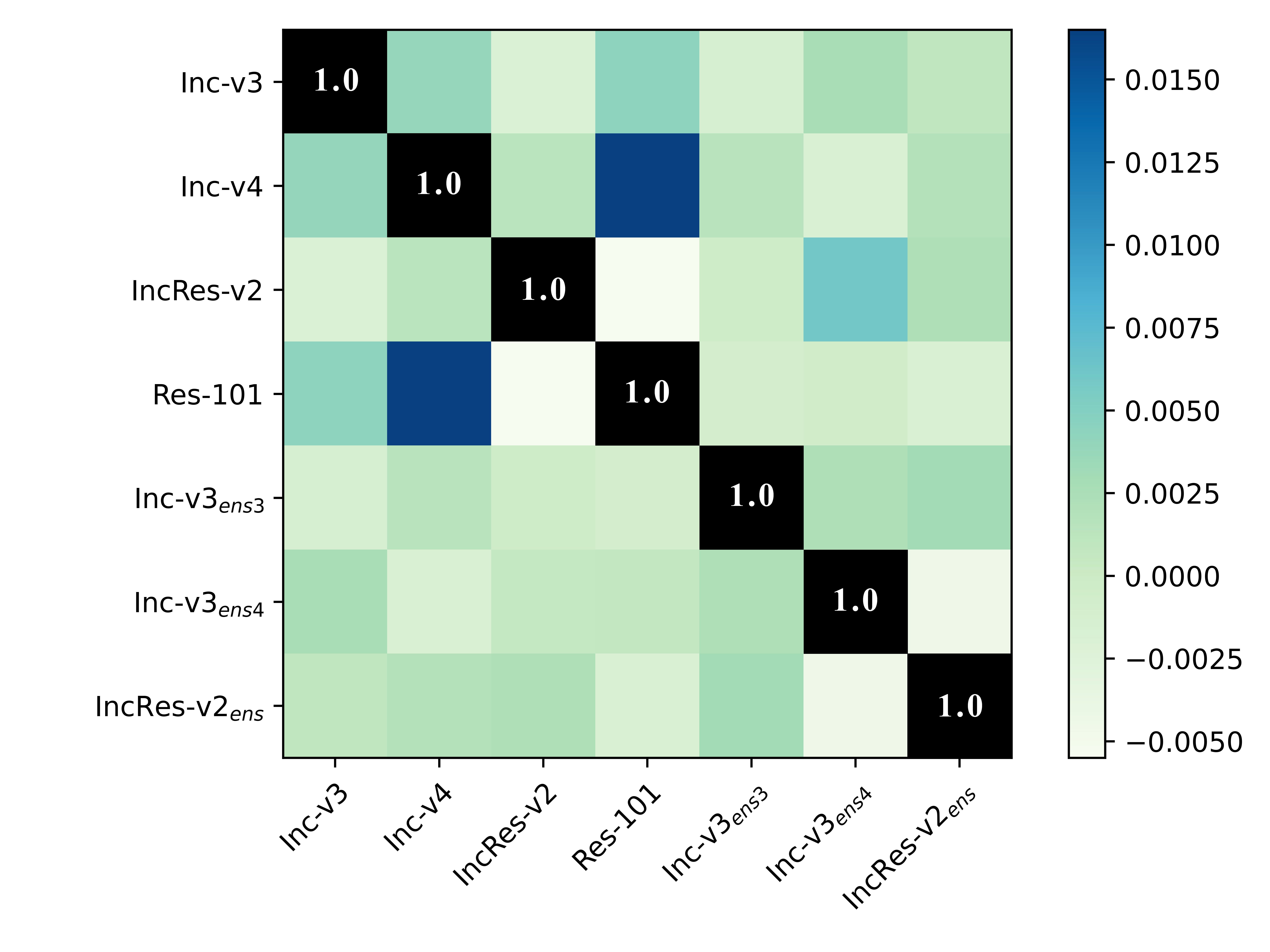}
\vspace{-1em}
\caption{The cosine similarity between the gradients (processed by sign function) of a sampled image on different models.}

\label{fig:cos}
\end{figure}

As demonstrated in Figure~\ref{fig:cos}, the cosine similarity between the update direction of a sampled image on different models is extremely low,
indicating there exists a considerable gap in the optimization direction among these models (See model details in Section \ref{section:setup}). 
We argue that simply fusing the predictions/logits/losses of the models but ignoring the variance of the gradients on different models would lead to a suboptimal result, and limit the performance of ensemble attacks.

\subsection{Stochastic Variance Reduced Ensemble Attack}
\label{section:methodology-3}

In previous works, Lin \etal and Wang \etal~\cite{lin2020nesterov, VT} analogize the process of the adversarial example generation to the process of neural network training, of which the white-box model is analogized to the training data and the black-box model is analogized to the test data. Hence, the iterative optimization on crafting the adversarial example using the input image can be regarded as the parameter update of neural networks, and the transferability of the adversarial example is analogized to the generalization of models. 
 
In this work, we treat the iterative ensemble attack as a stochastic gradient descent optimization process, in which at each iteration, the attacker always chooses the batch of the ensemble models for update. 
During the course of the adversarial example generation, the gradient variance on different models may lead to poor local optima. 
Hence, we aim to reduce the gradient variance so as to stabilize the gradient update direction, making the induced gradient be generalized better to other possible models. 
\par 

    


Inspired by the stochastic variance reduced gradient (SVRG) method~\cite{johnson2013accelerating} designed for stochastic optimization, we propose a stochastic variance reduced ensemble attack method to address the gradient variance of the models so as to take full advantage of the ensemble attack. 
The basic idea of SVRG is to reduce the inherent variance of Stochastic Gradient Descent (SGD) using predictive variance reduction, while we aim to reduce the inherent gradient variance of multiple models.
The integration of \name with MI-FGSM~\cite{dong2018boosting}, \name-MI-FGSM, is summarized in Algorithm~\ref{alg:svr_ema_ifgsm}.

\begin{algorithm}[t]

\caption{The \name-MI-FGSM attack algorithm}
\label{alg:svr_ema_ifgsm}
\begin{algorithmic}[1]
\REQUIRE A benign example $\mathbf x$ and its label $y$, a set of $K$ surrogate models and the corresponding losses $\{J_1,\ldots,J_{K}\}$, an ensemble loss $J$ chosen from $\{Eq.(\ref{eq:ens_on_predictions}),Eq.(\ref{eq:ens_on_logits}),Eq.(\ref{eq:ens_on_loss})\}$
\REQUIRE The perturbation bound $\epsilon$, number of iterations $T$, internal update frequency $M$, internal step size $\beta$, decay factor $\mu_1$,  internal decay factor $\mu_2$ 

\ENSURE An adversarial example $\bm{x}^{adv}$ that fulfills $\|\bm{x}^{adv} - \bm{x}\|_{\infty} \leq \epsilon$

\STATE $\alpha = \epsilon / T$; $\bm{G}_0 = 0$;
\STATE Initialize $\bm{x}_0^{adv} = \bm{x}$;

\FOR {$t = 0$ to $T-1$}


\STATE \textcolor{orange}{\# Calculate the gradient of the ensemble model} \\
\STATE Get the loss of the ensemble model $J(\bm{x}_t^{adv},y)$;\\
\STATE Calculate the gradient of the ensemble model $\bm{g}_{t}^{ens}$:\\ \qquad \qquad \qquad $\bm{g}_{t}^{ens} = \frac{1}{K} \nabla_{\bm{x}}J(\bm{x}_t^{adv},y)$;
\\
\STATE\textcolor{orange}{\# Stochastic variance reduction via $M$ updates}
\STATE Initialize $\bm{\tilde{x}}_0 = \bm{x}_t^{adv}$;$\bm{\tilde{G}}_0 = 0$
\FOR {$m = 0$ to $M-1$}
\STATE Randomly pick a model index $k \in \{1,\ldots,K\}$ 
\STATE Get the corresponding loss $J_{k} \in \{J_1,\ldots,J_{K}\}$ 
\STATE $\bm{\tilde{g}}_m = \nabla_{\bm{x}}J_{k}(\bm{\tilde{x}}_{m},y) - (\nabla_{\bm{x}}J_{k}(\bm{x}_{t}^{adv},y) - \bm{g}_{t}^{ens})$

\STATE \textcolor{orange}{\# Update the inner gradient by momentum}

\STATE $\bm{\tilde{G}}_{m+1} = \mu_2 \cdot \bm{\tilde{G}}_m + \bm{\tilde{g}}_m$
\STATE \textcolor{orange}{\# Update the inner adversarial example}
\STATE Update  $\bm{\tilde{x}}_{m+1} = {\rm Clip}_{x}^{\epsilon }\{\bm{\tilde{x}}_{m}+\beta\cdot {\rm sign}(\bm{\tilde{G}}_{m+1})\}$
\ENDFOR
\STATE \textcolor{orange}{\# Update the outer gradient by momentum}

\STATE $\bm{G}_{t+1} = \mu_1 \cdot \bm{G}_t + \bm{\tilde{G}}_{M}$
\STATE \textcolor{orange}{\# Update the outer adversarial example}
\STATE $\bm{x}_{t+1}^{adv} = {\rm Clip}_{x}^{\epsilon }\{{x}_{t}^{adv}+\alpha\cdot {\rm sign}(\bm{G}_{t+1})\}$

\ENDFOR

\STATE {\bfseries return} $\bm{x}^{adv} = \bm{x}_T^{adv}$
\end{algorithmic}
\end{algorithm}

Denote the traditional model ensemble attack method as Ens. 
The main difference of our method to Ens is that \name has an inner update loop, where \name obtains a variance reduced stochastic gradient by $M$ updates. Specifically, we first obtain the gradient of the multiple models, $\bm{g}^{ens}$, by one pass over the models and maintain this value during $M$ inner iterations. Then, we randomly pick a model from the ensemble models, obtain the stochastic inner gradient after variance reduction $\bm{\tilde{g}}_m$, and update the inner adversarial example using 
the accumulate gradients of $\bm{\tilde{g}}_m$. 
In the end, we update the outer gradient using the accumulate gradient 
of the last inner loop.    
As $\bm{\tilde{g}}_m$ is the unbiased estimate of the gradient of $\bm{g}_{m}^{ens}$, $(\nabla_{\bm{x}}J_{k}(\bm{x}_{t}^{adv},y) - \bm{g}_{t}^{ens})$ helps to reduce the gradient on different models. 

In a nutshell, the existing Ens method directly uses the average gradient of the ensemble models $\bm{g}^{ens}$ to update the adversarial example, while \name uses the stochastic variance reduced gradient $\bm{\tilde{g}}$ to update the adversarial example. 
Theoretically, \name can be easily integrated with other iterative gradient-based attack methods. 
\Eg I-FGSM~\cite{Goodfellow2014Explaining}, MI~\cite{dong2018boosting}, DI~\cite{dong2019evading},  TI~\cite{dong2019evading}, SI~\cite{lin2020nesterov} can be integrated with \name using the same technique in inner loop and outer loop. But in \name-I-FGSM, we accumulate gradients in inner loop to have a better transferbility.

Compared with existing optimization-based methods of enhancing the attack transferability, our method is from a different perspective. Existing works mainly focus on the optimization along the iterative process. For instance,  MI-FGSM~\cite{dong2018boosting} and NI-FGSM~\cite{lin2020nesterov} aim to accelerate the convergence, while VT~\cite{VT} aims to tune the current gradient using the variance of the gradient at the previous iteration for a single model. In contrast, our method seeks to reduce the variance of the gradient caused by various models in the ensemble attack.

\section{Experiments}
\label{section:experiments}
This section first introduces the experimental setup,  
then reports the attack success rate on normally trained models and defense models, showing that \name outperforms Ens significantly for black-box attacks.  
We further demonstrate that \name increases the average loss on black-box models by a large margin. In the end, we perform ablation studies to manifest the effectiveness of the key parameters in \name.

\subsection{Experimental Setup}
\label{section:setup}
\textbf{Dataset.} We conduct experiments on an ImageNet-compatible dataset\footnote{\url{https://github.com/cleverhans-lab/cleverhans/tree/master/cleverhans_v3.1.0/examples/nips17_adversarial_competition/dataset}} which is comprised of 1,000 images and is widely used in recent
FGSM-based attacks~\cite{dong2019evading, gao2020patch}.
\par

\textbf{Networks.} We consider four normally trained networks, \ie, Inception-v3 (Inc-v3)~\cite{incv3}, Inception-v4 (Inc-v4), Resnet-v2-152 (Res-152)~\cite{incv4}, and Inception-Resnet-v2 (IncRes-v2)~\cite{res101}. For adversarially trained models, we consider Inc-v3${\rm _{ens3}}$, Inc-v3${\rm _{ens4}}$ and IncRes-v2${\rm _{ens}}$~\cite{ensmodel}. Besides, we consider nine defense models which are shown to be robust against black-box attacks, including the top-3 defense methods in the NIPS competition: HGD~\cite{defense-HGD}, R\&P~\cite{defense-RP},
NIPS-r3~\footnote{\url{https://github.com/anlthms/nips-2017/tree/master/mmd}} and six recently proposed defense methods: Bit-R~\cite{defense-Bit-red}, JPEG~\cite{defense-JPEG}, FD~\cite{defense-FD}, ComDefend~\cite{defense-ComDefend}, RS~\cite{defense-RS} and NRP~\cite{defense-NRP}.
\par

\textbf{Baselines.}
We compare the proposed \name with Ens based on the advanced gradient-based attacks, including I-FGSM~\cite{Goodfellow2014Explaining}, MI-FGSM~\cite{dong2018boosting}, TIM~\cite{dong2019evading}, TI-DIM~\cite{dong2019evading}, and SI-TI-DIM~\cite{lin2020nesterov}.
For Ens, we adopt the ensemble method that fuses the logits of difference models~\cite{dong2018boosting}, which is confirmed better than the ensemble on predictions or losses. In addition, we run the \name attack for 5 times with different random seeds and average the results to reduce the impact of randomness. 

\par
\textbf{Hyper-parameters.}
To align with the previous works~\cite{dong2018boosting, xie2019improving, dong2019evading,lin2020nesterov}, we set the maximum perturbation 
$\epsilon$ = 16/255. For I-FGSM, the number of iterations is 10, and the step size is $\alpha$ = 1.6. For MI-FGSM, we set the decay factor $\mu_1$ to 1.0. For TIM, we adopt the Gaussian kernel with size $7 \times 7$. For TI-DIM, the transformation probability $p$ is set to 0.5. For SI-TI-DIM, we set the number of copies $m$ to 5. For \name,  we set the internal update frequency $M$ to four times the number of ensemble models and the internal step size $\beta$ is set the same as $\alpha$, the internal decay factor $\mu_2$ is set to 1.0. 
\begin{table}[h]
\caption{The attack success rates (\%) of adversarial examples against the hold-out model. 
We study four normal models: Inc-v3, Inc-v4, IncRes-v2 and Res-101. 
For each model, the adversarial examples are crafted on an ensemble of the other three.
}
\vspace{-0.2cm}
\label{tab:normal}
\begin{center}
\resizebox{0.48\textwidth}{!}{
\begin{tabular}{c|c|c|c|c|c|c} 
\hline
   Base & Attack & Inc-v3 & Inc-v4 & IncRes-v2 & Res-101 & Average \\
\hline \hline
              \multirow{2}*{I-FGSM} & Ens & 77.30 & 66.70 & 58.50 & 48.80 &62.83
              \\
              & {SVRE} & \textbf{89.24} & \textbf{83.64} & \textbf{77.60} & \textbf{65.58}& \textbf{79.02}\\
\hline
              \multirow{2}*{MI-FGSM} & Ens &90.30 &86.60 & 82.20 &77.40  & 84.13\\
              & {SVRE} & \textbf{96.84} & \textbf{95.30} & \textbf{92.80} &\textbf{89.40}& \textbf{93.59}\\
           \hline
              \multirow{2}*{TIM} & Ens  & 91.70& 88.70 &84.30 & 79.20  &85.98\\
              & {SVRE} &\textbf{96.10}& \textbf{93.66} & \textbf{90.18}& \textbf{85.36}&\textbf{91.33}\\
              \hline
              \multirow{2}*{TI-DIM} & Ens & 95.70 &94.10 &93.20 &90.10 & 93.28\\
              & {SVRE} & \textbf{97.78} & \textbf{96.86} &\textbf{95.92} & \textbf{93.98}&\textbf{96.14}\\
              \hline
              \multirow{2}*{SI-TI-DIM} & Ens & 97.60 &97.60  &97.20 & 95.90 &97.08\\
              & {SVRE} & \textbf{98.80}& \textbf{98.88} & \textbf{97.90} & \textbf{97.82} &\textbf{98.35}\\
\hline
\end{tabular}}
\end{center}
\vspace{-0.2cm}
\end{table}

\subsection{Attack Normally Trained Models}
We first compare the performance of our method on the normally trained models, including Inc-v3, Inc-v4, Res-152 and IncRes-v2. Specifically, we keep one model as the hold-out black-box model and generate adversarial examples on an ensemble of the other three models by Ens and \name integrated with various base methods.

Table ~\ref{tab:normal} shows the attack performance on the hold out model. 
\name outperforms Ens across all the test models. 
The average improvement of \name over Ens on the base attack of I-FGSM is significant at 16.19\%.
Even on the advanced attack methods, MI-FGSM, TIM, DIM and SI-TI-DIM, the average improvements of \name over Ens are still considerable, which are 9.46\%, 5.35\%, 2.86\% and 1.27\%, respectively. 
The results demonstrate that \name can effectively improve the transferability of adversarial examples on normally trained models. 



\label{section:normal}

\subsection{Attack Advanced Defense Models}\label{section:defense}

\begin{table*}[th]
\caption{The black-box attack success rates (\%) against three adversarially trained models. The adversarial examples are generated on the ensemble models, \ie, Inc-v3, Inc-v4, IncRes-v2 and Res-101.
}
\vspace{-1.3ex}
\label{tab:advtrain}
\begin{center}
\resizebox{\textwidth}{!}{
\begin{tabular}{c|c|c|c|c|c|c|c|c|c} 
\hline

   \multirow{2}*{Base} & \multirow{2}*{Attack} & \multicolumn{4}{c}{White-box setting} &  \multicolumn{4}{|c}{Black-box setting}\\
   \cline{3-6} \cline{7-10} 
        & & Inc-v3 & Inc-v4 & IncRes-v2 & Res-101 & Inc-v3${\rm _{ens3}}$& Inc-v3${\rm _{ens4}}$ &IncRes-v2${\rm _{ens}}$ & Average \\

\hline \hline
              \multirow{2}*{I-FGSM} & Ens &100.00 & 100.00  &  99.60& 99.80 & 27.10  &24.50  &15.70 &22.43\\
              & {SVRE} & 99.80  &99.60  &99.38  &99.58  &\textbf{40.08} &\textbf{37.30} &\textbf{24.76}    & \textbf{34.05}\\
\hline
             \multirow{2}*{MI-FGSM} & Ens  &99.90   &  99.90 & 99.70& 99.50  &50.50  &49.30 & 32.30  &44.03\\
              & {SVRE} & 99.96  &99.96  &99.86  &99.82   & \textbf{64.54} & \textbf{59.02} & \textbf{39.08} &\textbf{54.21} \\
           
           \hline
              \multirow{2}*{TIM} & Ens  & 99.80&  99.70 & 99.40 &99.20  &73.50   &68.10  & 59.70 & 67.10\\
              & {SVRE} &  99.84 & 99.90 & 99.80 & 99.70  &\textbf{87.88}  &\textbf{85.62} &\textbf{79.70} &\textbf{84.40} \\
          
              \hline
              \multirow{2}*{TI-DIM} & Ens & 99.50 & 99.40 &99.00 &98.70 & 87.40 &84.30 &77.60 & 83.10 \\
              & {SVRE} & 99.86 &99.80 & 99.68 &99.34 & \textbf{95.32} & \textbf{93.66} &\textbf{90.08} & \textbf{93.02} \\
              \hline
              \multirow{2}*{SI-TI-DIM} & Ens &99.70&99.40 &99.30 &99.40 &95.60 &95.10 &92.40 &94.37\\
              & {SVRE} & 99.98 &99.96 &99.90 &99.80 & \textbf{98.56} &\textbf{97.78} &\textbf{95.80} &\textbf{97.38} \\
              
\hline
\end{tabular}}

\end{center}

\end{table*}

\begin{table*}[tp]
\caption{The black-box attack success rates (\%) against nine models with advanced defense mechanism.
}
\vspace{-0.2cm}
\label{tab:defense}
\begin{center}
\resizebox{\textwidth}{!}{
\begin{tabular}{c|c|c|c|c|c|c|c|c|c|c|c} 
\hline
   Base & Attack & HGD & R\&P & NIPS-r3 & Bit-R &  JPEG & FD & ComDefend & RS & NRP  & Average \\

\hline \hline
  
    \multirow{2}*{I-FGSM} & Ens      & 27.00 & 15.20 &18.90  &26.00
    &41.80 &37.10 &56.00  &25.20  &17.30 &29.39\\
    & {SVRE}  & \textbf{45.48} & \textbf{25.02} &\textbf{34.10}  &\textbf{30.96}
    &\textbf{62.06} &\textbf{50.42} &\textbf{66.98}&\textbf{26.98}  &\textbf{21.60} & \textbf{40.40}\\
    \hline
    \multirow{2}*{MI-FGSM} & Ens     & 41.30 & 33.00 &44.60  &39.70
    &75.90 &62.80 & 77.50 &36.90  &27.30 &48.78\\
    & {SVRE} & \textbf{44.06} & \textbf{40.72} &\textbf{59.54}  &\textbf{43.42}
    &\textbf{89.06} &\textbf{73.28} &\textbf{86.60}&\textbf{39.12}  &\textbf{28.46} &\textbf{56.03}\\
    \hline       
         
    \multirow{2}*{TIM} & Ens     & 72.50 & 60.50 &67.20  &49.30
    &82.60 &74.80 &85.10  &47.80 &37.60 &64.16\\
    & {SVRE} & \textbf{87.10} & \textbf{80.16} &\textbf{83.84}  &\textbf{62.26}
    &\textbf{91.96} &\textbf{83.96} &\textbf{92.22}&\textbf{62.46}  &\textbf{52.24} &\textbf{77.36}\\
             
              \hline
    \multirow{2}*{TI-DIM} & Ens 
               & 87.40 & 81.20  &85.70  &63.00   &91.70 &84.30 &91.90  &57.90  &49.80 &76.99\\
    & {SVRE} 
               & \textbf{94.86} & \textbf{91.92} &\textbf{93.22}  &\textbf{72.88}   &\textbf{96.48} &\textbf{90.76} &\textbf{95.98}  &\textbf{73.60}  &\textbf{65.38} &\textbf{86.12}\\
               \hline
    \multirow{2}*{SI-TI-DIM} & Ens 
               & 95.70 & 93.20 &94.10  &82.70   &96.70 &93.30 & 97.90 &78.00  &76.80 &89.82\\
    & {SVRE} 
               & \textbf{97.70} & \textbf{96.12} &\textbf{97.48}  &\textbf{86.64}   &\textbf{98.54} &\textbf{95.60} &\textbf{99.06}&\textbf{85.72} &\textbf{85.44} &\textbf{93.59}\\ 
              
\hline
\end{tabular}}
\end{center}
\end{table*}

To further validate the efficacy of our method in practice, 
we continue to evaluate \name on models with various advanced defenses. 
Specifically, we craft the adversarial examples on the ensemble of four normally trained models, \ie, Inc-v3, Inc-v4, Res-15 and IncRes-v2, and test the transferability of the crafted adversaries on defense models. \par

We first evaluate the transferability of the adversaries on three adversarially trained models, Inc-v3${\rm _{ens3}}$, Inc-v3${\rm _{ens4}}$ and IncRes-v2${\rm _{ens}}$. The results are shown in Table ~\ref{tab:advtrain}.
We see that \name outperforms Ens on the black-box attack of each adversarially trained models by a large margin. 
Among the base methods that the ensemble attacks integrate with, \name exhibits the highest improvement on TIM, as \name-TIM yields a 17.30\% higher average attack success rate than Ens-TIM. Besides, \name also performs well in the white-box setting, and can slightly improve the white-box attack performance in most cases.

In addition to the adversarially trained models, we also evaluate the crafted examples on nine models with advanced defense mechanisms. 
The results are shown in Table ~\ref{tab:defense}. 
\name outperforms Ens by a clear margin across all the comparisons. 
The strongest version of our method, \name integrated with SI-TI-DIM, 
can achieve an average attack success rate of $93.59\%$ 
on these defense models in the black-box setting, which raises a new security issue for the robustness of deep learning models.


\subsection{Comparison on Loss}
\label{section:Comparison}
\begin{figure*}[tb!]
\centering

\begin{subfigure}{.5\textwidth}
        \centering 
        \includegraphics[width=\linewidth]{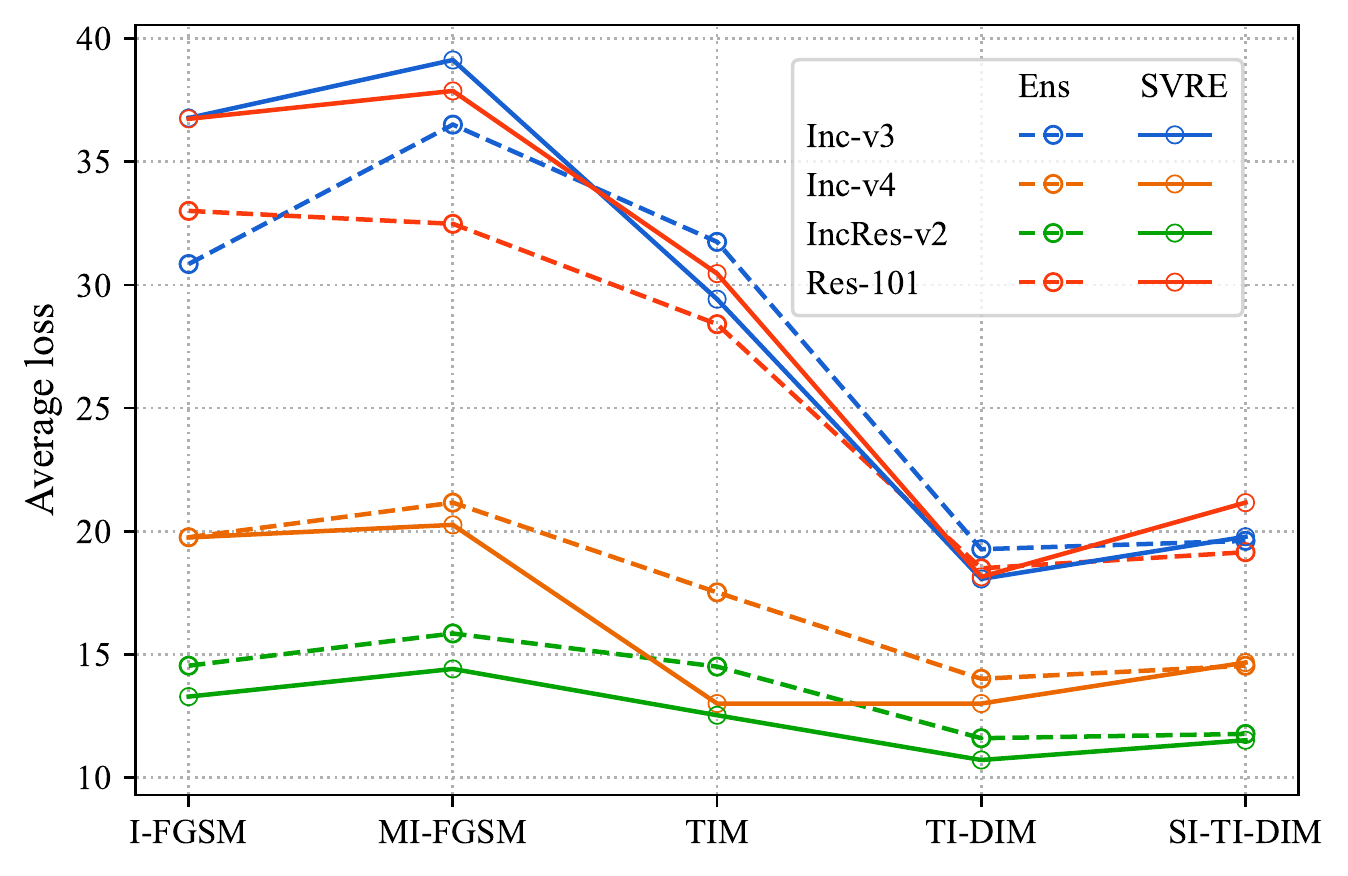}
        \vspace{-0.5cm}
        
        \caption{White-box setting}
    \end{subfigure}%
    \begin{subfigure}{.5\textwidth} 
        \centering 
        \includegraphics[width=\linewidth]{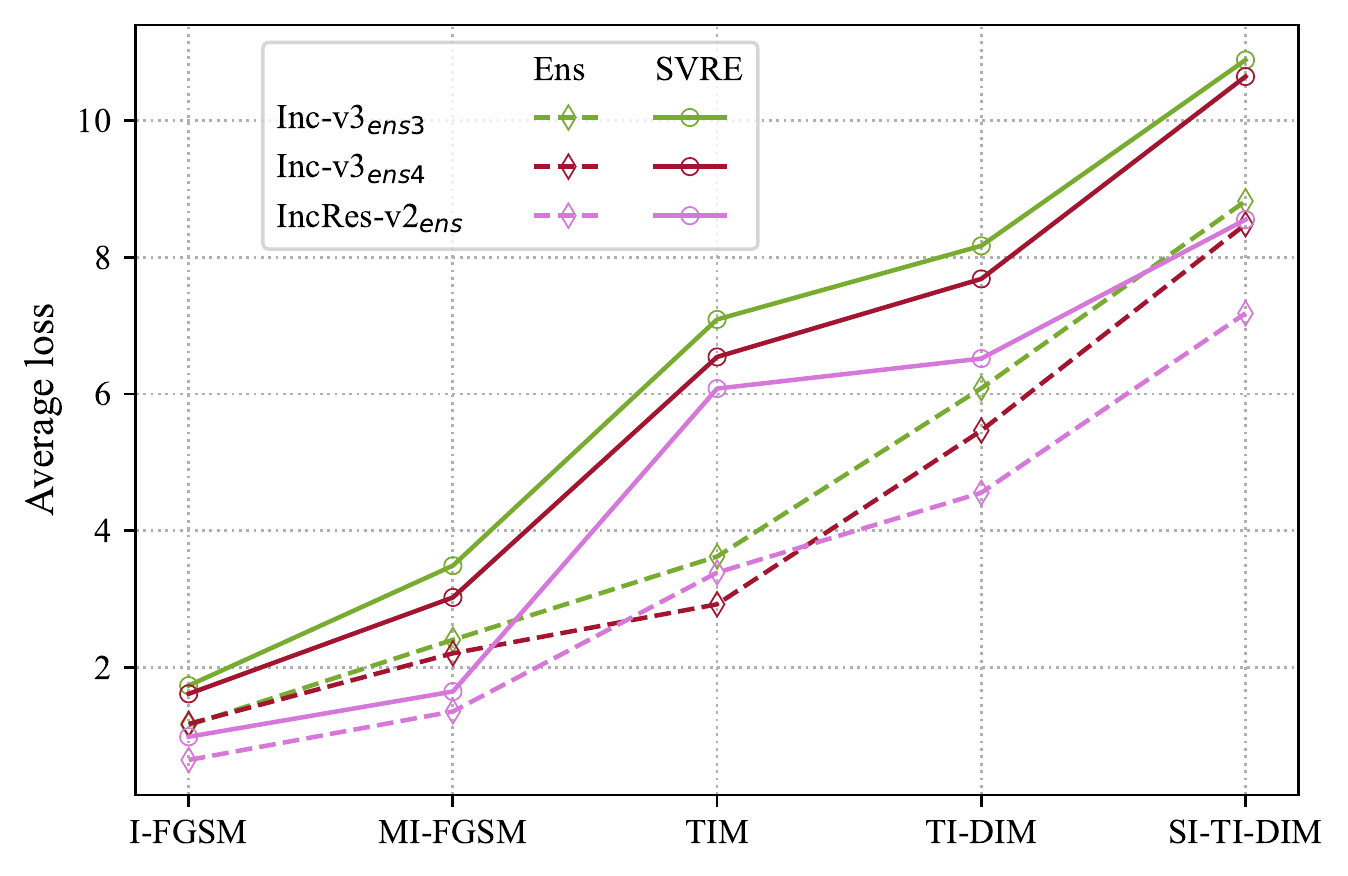}
        \vspace{-0.5cm}
        \caption{Black-box setting}
    \end{subfigure}%

\caption{The average loss on seven models against Ens and \name integrated with five attacks, respectively.}\label{fig:loss}
\end{figure*}

The above experiments have demonstrated that \name significantly improves the attack success rate of adversarial attacks. To provide intuitive evidence that \name can effectively boost the transferability of adversarial examples, we average the loss over the adversarial images generated in Section \ref{section:defense} on four white-box models and three black-box models respectively,
and depict the improvement curve for the average loss in Figure \ref{fig:loss}. The loss can indirectly reflect the adversarial efficacy. 
A higher loss indicates a stronger adversarial intensity, and a higher loss on the black-box model indicates a stronger transferability.
\par 

We can see in Figure \ref{fig:loss} (b) that \name increases the average loss over Ens on black-box models remarkably. 
In terms of the white-box setting in Figure \ref{fig:loss} (a), \name and Ens are comparative, indicating that the improvement of \name in transferability is not based on the premise of sacrificing the performance of white-box attacks.

\begin{figure*}[t!]
 \begin{subfigure}{.333\textwidth}
        \centering 
        \includegraphics[width=0.78\linewidth]{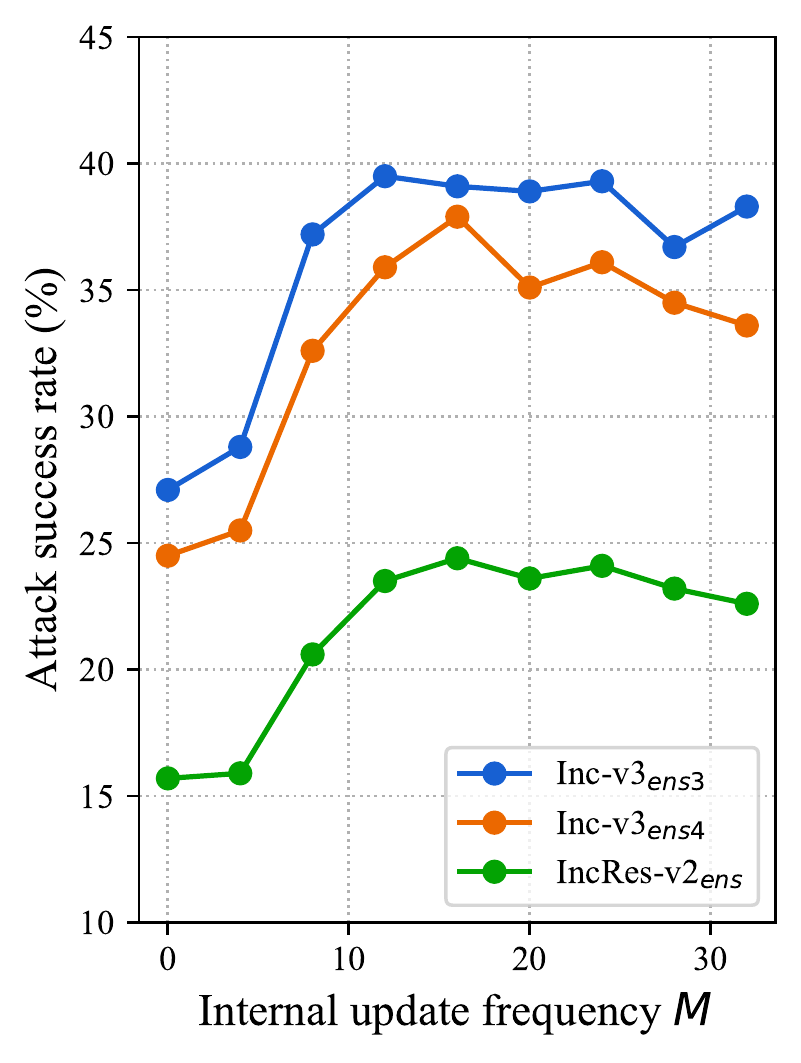}

        \caption{\name-I-FGSM}
    \end{subfigure}%
    \begin{subfigure}{.333\textwidth} 
        \centering 
        \includegraphics[width=0.78\linewidth]{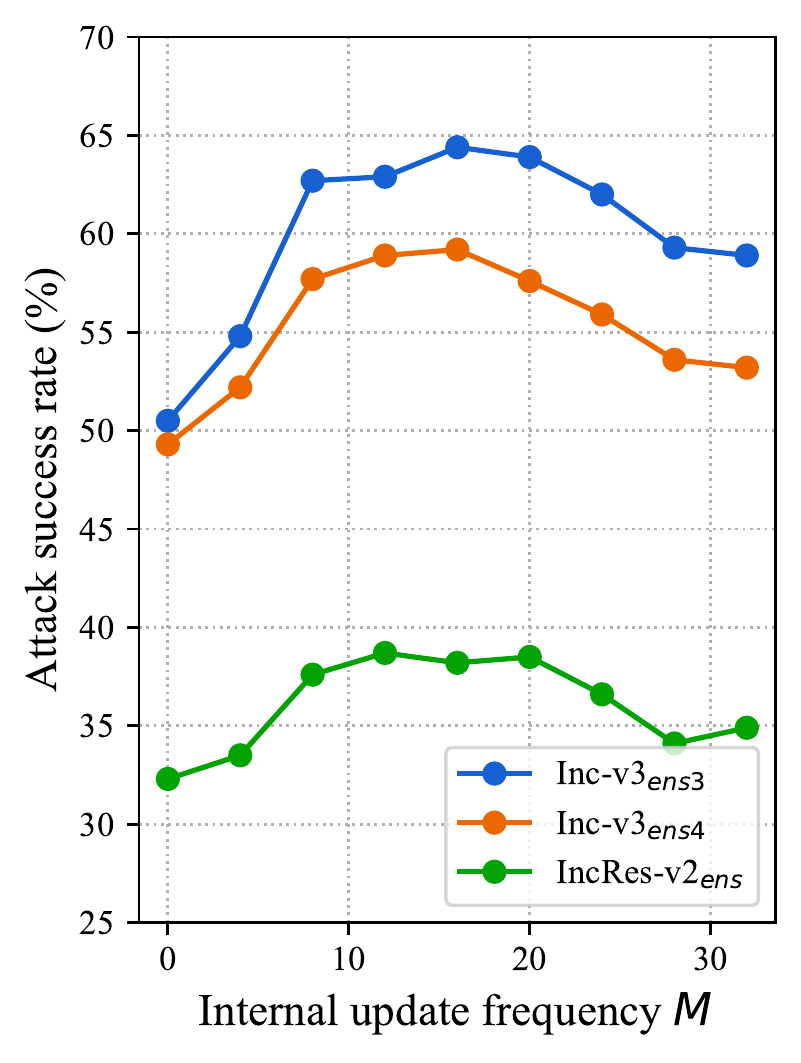}
       
        \caption{\name-MI-FGSM}
    \end{subfigure}%
    \begin{subfigure}{.334\textwidth}
        \centering 
        \includegraphics[width=0.78\linewidth]{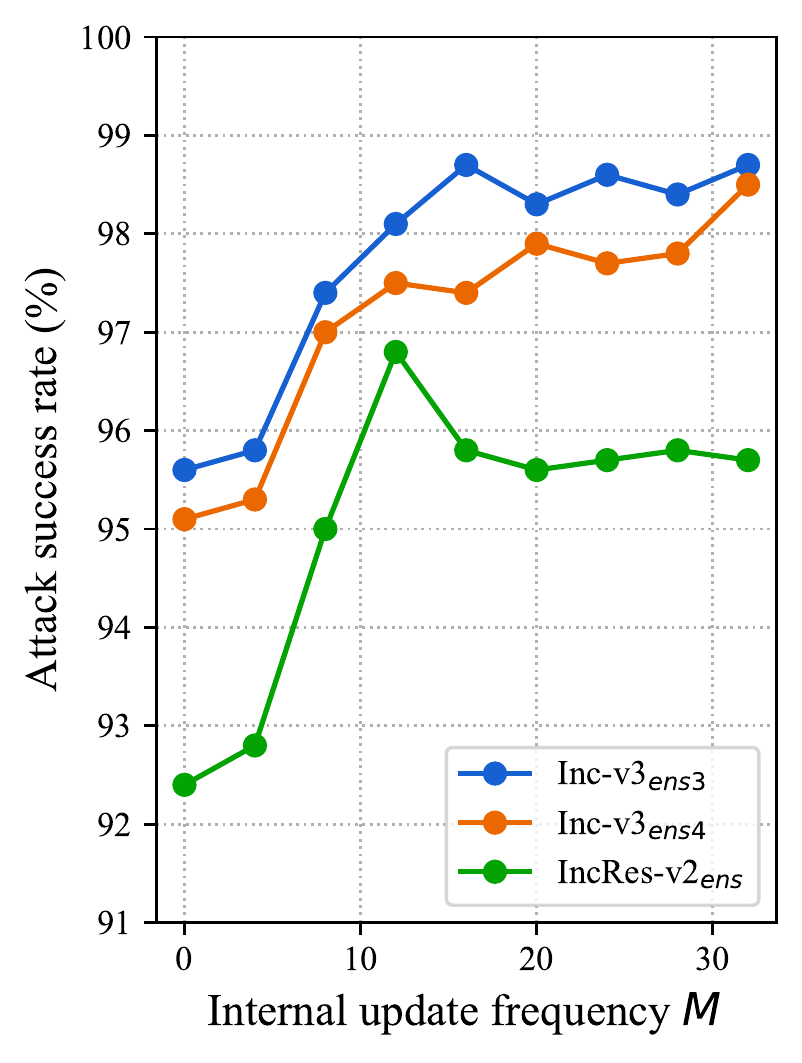}
        
         \caption{\name-SI-TI-DIM}
    \end{subfigure}
    \caption{The attack success rate (\%) of \name integrated with I-FGSM, MI-FGSM and SI-TI-DIM. It degenerates to the integration with Ens when $M = 0$.}
    \label{fig:svrgit}
\end{figure*}

\begin{figure*}[tb!]
 \begin{subfigure}{.333\textwidth}
        \centering 
        \includegraphics[width=0.78\linewidth]{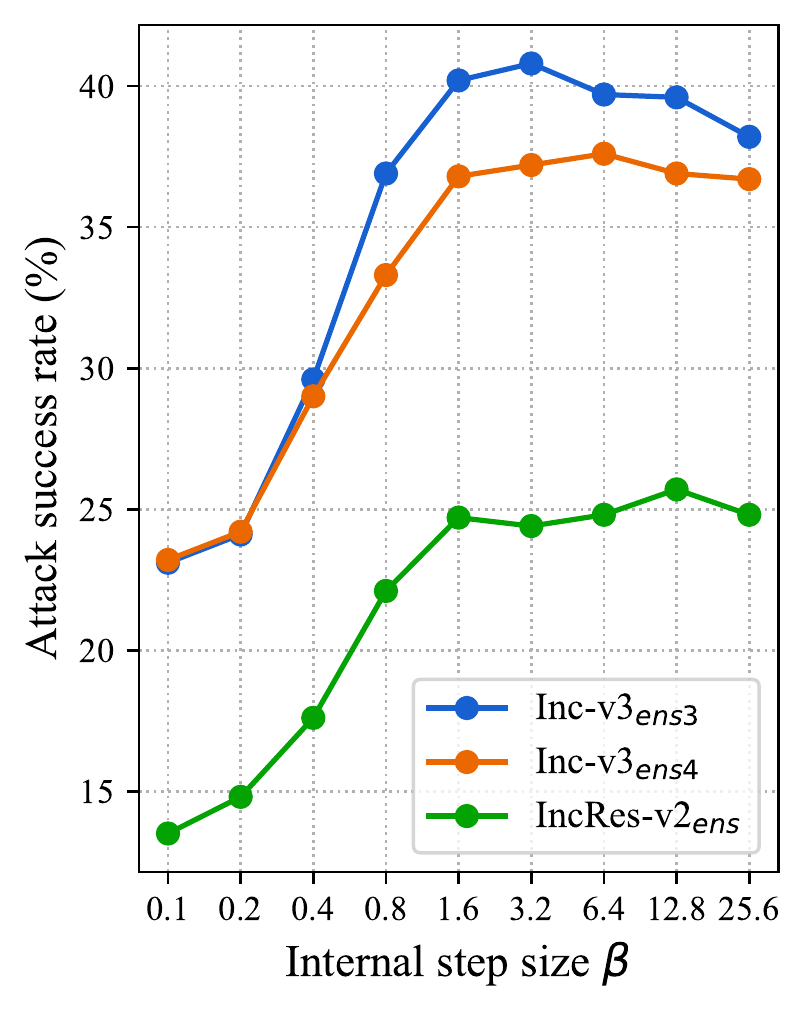}
        
        \caption{\name-I-FGSM}
    \end{subfigure}%
    \begin{subfigure}{.333\textwidth} 
        \centering 
        \includegraphics[width=0.78\linewidth]{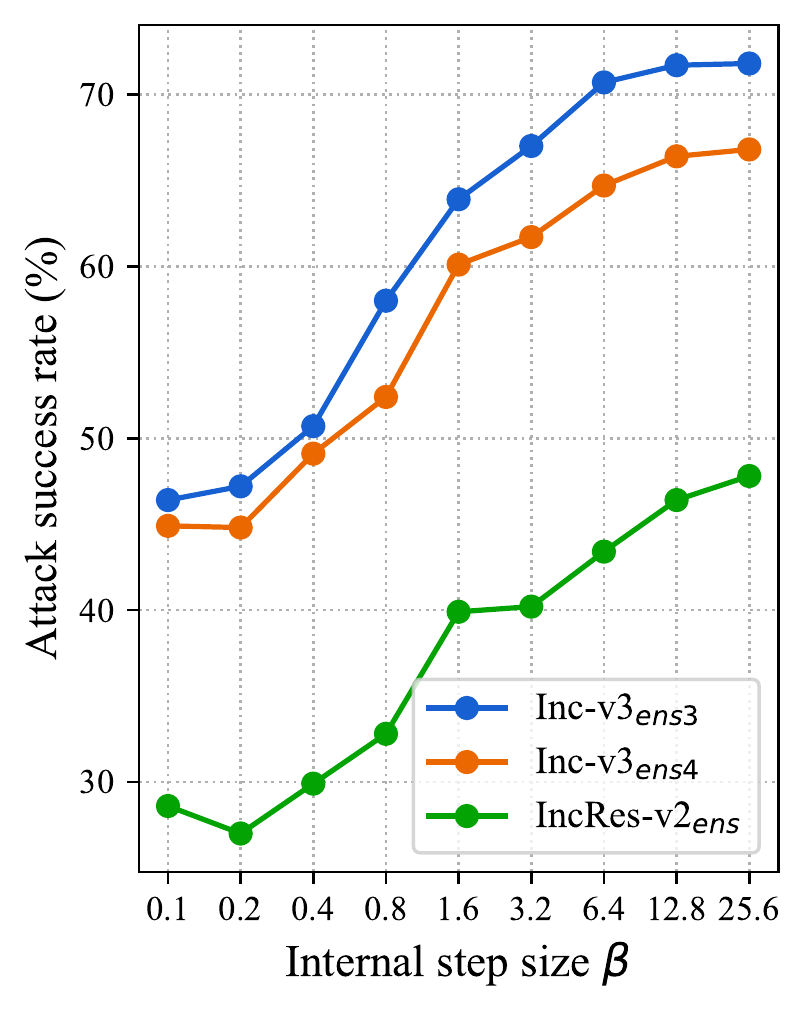}
        \caption{\name-MI-FGSM}
    \end{subfigure}%
    \begin{subfigure}{.34\textwidth}
        \centering 
        \includegraphics[width=0.78\linewidth]{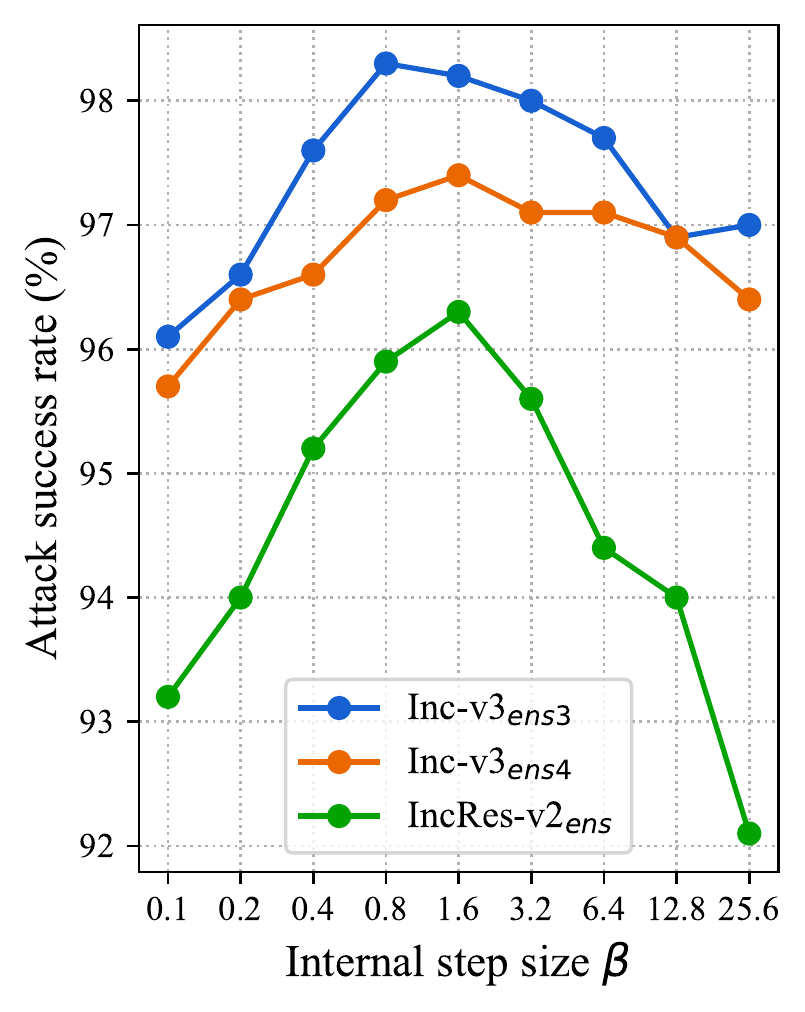}
         \caption{\name-SI-TI-DIM}
    \end{subfigure}

    \caption{The attack success rate (\%) of I-FGSM, MI-FGSM and SI-TI-DIM after integrated with \name on different internal step size $\beta$.}

    \label{fig:stepsize}
\end{figure*}

\subsection{Ablation Study on Hyper-parameters}
\label{section:ablation}

In this subsection, we conduct a series of ablation experiments to study the impact of the parameters in \name. 
Here we attack the ensemble of Inc-v3, Inc-v4, Res-152 and IncRes-v2 and test the transferability of the adversaries on the adversarially trained models Inc-v3${\rm _{ens3}}$, Inc-v3${\rm _{ens4}}$ and IncRes-v2${\rm _{ens}}$, as the setting in Section \ref{section:normal}.
 
\begin{figure*}[tb]
\centering
\includegraphics[width=.65\linewidth]{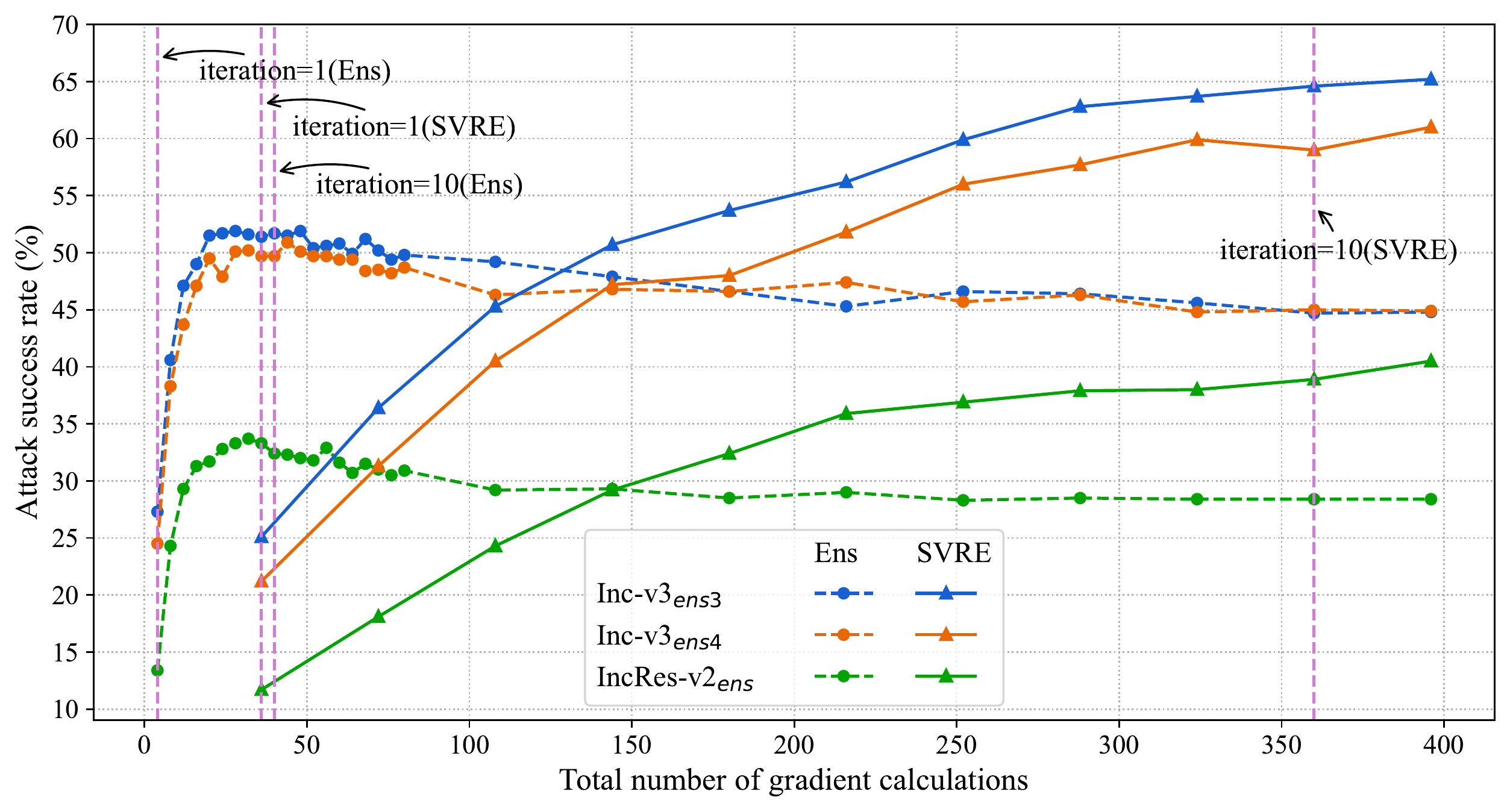}

\caption{The attack success rate (\%) of \name-MI-FGSM and Ens-MI-FGSM for different total number of gradient calculations.}
\label{fig:iteration}
\end{figure*}

\textbf{On the internal update frequency $M$.} 
We first analyze the effectiveness of the internal update frequency $M$ on the attack success rate of \name. We integrate I-FGSM, MI-FGSM and SI-MI-DIM attacks with \name, respectively, and range the internal update frequency $M$ from 0 to 32 with a granularity of 4. Note that if $M = 0$, \name trivially degenerates to the normal ensemble method of Ens. Since the attack success rate in the white-box setting is close to 100\%, we only show the results for black-box attacks,
as shown in Figure \ref{fig:svrgit}. A first glance shows that our \name has achieved an impressive improvement over Ens ($M = 0$). As the number of iterations increases, the attack success rate increases and reaches the peak at about $M = 16$. We also observe from the convex curve that either too many iterations or too few iterations may cause the adversarial examples to overfit the current model and harm the attack transferability. 

\par
\textbf{On the internal step size $\beta$.}
The internal step size $\beta$ plays a crucial role in improving the attack success rate, as it determines the extent of the data point update in each inner loop. Similarly, we perform \name integrated with I-FGSM, MI-FGSM and SI-MI-DIM respectively, fix $\alpha = 1.6$ and let $\beta$ ranges from 0.1 doubled to 25.6. As shown in Figure \ref{fig:stepsize}, the performance of \name varies with the step size, and the best step size varies for different methods. 
For a fair comparison, we did not deliberately set different best parameters for each method but choose 
$\beta = 1.6$. For practical applications, the best step size can be adopted for a specific attack to obtain a higher performance. 

\textbf{On the number of iterations $T$.} 
For the same number of iterations, \name has more gradient calculations due to its inner loop. To show that the gains of \name is not simply from increasing the number of gradient calculations, we perform additional analysis on the total number of iterations. 
Taking the internal update frequency $M = 16$ and the number of ensemble models $K = 4$ as an example, each iteration requires 4 queries of the model in Ens, while for \name, the inner loop requires $16 \times 2 = 32$ additional queries. The overall number of queries for \name is 9 times that of Ens.
Then, what if we increase the number of iterations for other methods? 
One can observe from Figure \ref{fig:iteration} that the attack success rate of Ens-MI-FGSM against black-box models gradually decays with the increment on the total number of gradient calculations, and there is a big gap even when the total number reaches 360. 
This experiment demonstrates that simply increasing the number of iterations on Ens could not gain the high attack performance of SVRE.

\section{Conclusion}
\label{section:conclusion}
In this work, we propose a novel method called the stochastic variance reduced ensemble (\name) attack to enhance the transferability of the crafted adversarial examples. Different from the existing model ensemble attacks that simply fuse the outputs of multiple models evenly, the proposed \name takes the gradient variance of different models into account and reduces the variance to stabilize the gradient update on ensemble attacks. 
In this way, \name can craft adversarial examples with higher transferability for other possible models. 
Extensive experiments demonstrate that \name outperforms the vanilla model ensemble attack in the black-box setting significantly, at the same time \name keeps roughly the same performance in the white-box setting.  

Compared with broad investigations on the gradient optimization or input transformation attacks, the ensemble attacks are less explored in previous studies. Our work could shed light on the great potential of boosting the adversarial transferability through a better design on the ensemble methods. In future work, we wish our work could inspire more in-depth works in this direction for ensemble attacks. 

\section*{Acknowledgements}
This work is supported by National Natural Science Foundation of China (62076105) and  Hubei International Cooperation Foundation of China (2021EHB011).

{
    \small
    \bibliographystyle{ieee_fullname}
    \bibliography{macros,main}
}

\appendix

\setcounter{page}{1}

\twocolumn[
\centering
\Large
\textbf{Appendix} \\
\vspace{3.0em}
] 
\appendix

\section{Analysis on Training Time}
In comparison to Ens, our \name introduces an extra loop, which brings additional costs for crafting adversarial examples. The complexity is proportional to the total number of queries, so SVRE is $(2M+n)/n$ times of Ens where $M$ ($M = 16$) is the internal update frequency and $n$ ($n = 4$) is the number of ensemble models.
Note that other efforts for improving the adversarial transferability also introduce additional costs. 
\Eg, SIM~\cite{lin2020nesterov} makes $m = 5$ copies of the input for querying, VT~\cite{VT} samples $N = 20$ neighborhoods for variance tuning, and Admix~\cite{DBLP:journals/corr/abs-2102-00436} randomly sample $m_2 = 3$ images from other categories and copy each image for $m_1 = 5$ times.
In the light of the improved performance, the additional time cost is acceptable. 

\section{SVRE with other Advanced Method}
To show how \name compares to the state-of-the-art black-box adversarial attacks, 
we further incorporate \name with Admix~\cite{DBLP:journals/corr/abs-2102-00436}, the most recent black-box attack method, and show how \name help promote its performance. 
Specifically, we use SVRE-Admix-TI-DIM and Ens-Admix-TI-DIM to generate adversarial examples, respectively,
on the ensemble of Inc-v3, Inc-v4, IncRes-v2 and Res-152,
and test on three adversarially trained models,
while the Admix-TIM-DIM base method crafts adversarial examples on the Inc-v3 model. 

The results are  summarized in Table \ref{tab:addition}. We can see that the ensemble attack of Ens-Admix-TI-DIM has considerably higher transferability than Admix-TI-DIM, and our SVRE-Admix-TI-DIM further promotes the performance. 

\section{Visualization on Crafted Examples}
\begin{figure}[t]
\centering
\includegraphics[width=1.0\linewidth]{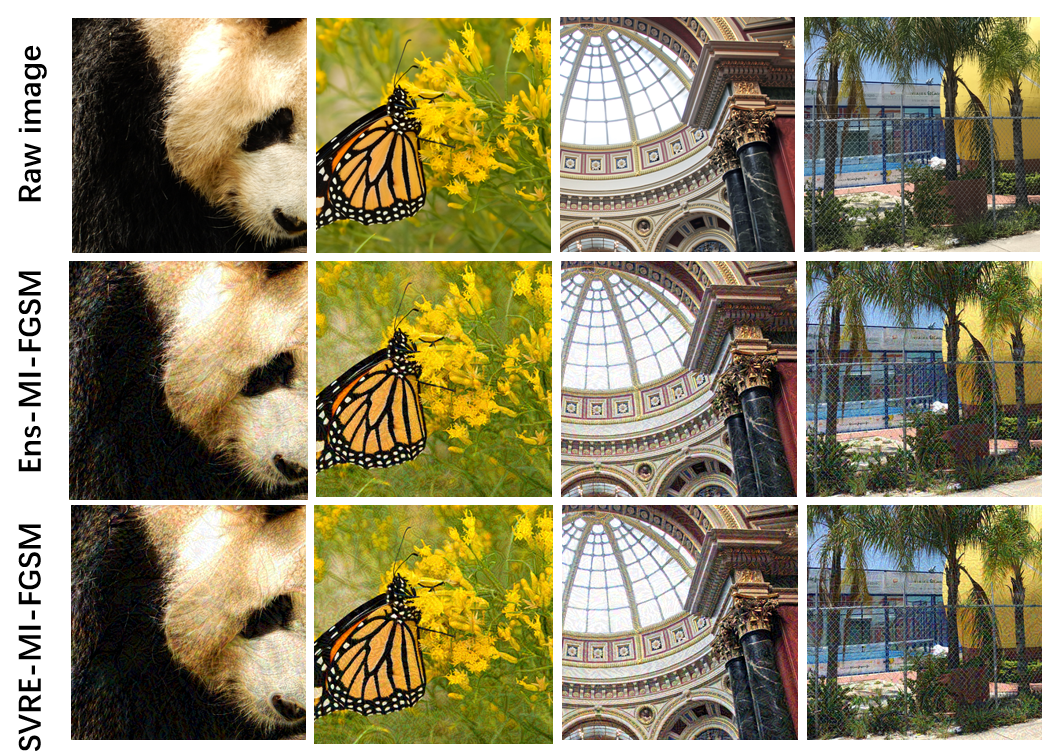}
\caption{Adversarial examples generated by Ens-MI-FGSM and \name-MI-FGSM, respectively. The adversarial examples are crafted on an ensemble of Inc-v3, Inc-v4, IncRes-v2 and Res-152 models.}

\label{fig:Visualizations}
\end{figure}

In Figure \ref{fig:Visualizations}, we visualize six randomly selected raw images and their corresponding adversarial examples crafted by Ens-MI-FGSM and \name-MI-FGSM, respectively. The adversarial examples are crafted on the ensemble of Inc-v3, Inc-v4, IncRes-v2 and Res-152 models. It shows that the adversarial examples crafted by our \name method are imperceptible to human eyes.

\begin{table}[t]
\caption{The black-box attack success rates (\%) against three adversarially trained models using Admix-TI-DIM as the base method. 
}
\vspace{-1em}
\label{tab:addition}
\begin{center}
\resizebox{0.48\textwidth}{!}{
\begin{tabular}{c|c|c|c|c} 
\hline
   Attack method & Inc-v3${\rm _{ens3}}$& Inc-v3${\rm _{ens4}}$ &IncRes-v2${\rm _{ens}}$ & Average\\
    \hline 
    Admix-TI-DIM & 80.8 & 78.9 & 63.2 & 74.3\\ 
    Ens-Admix-TI-DIM & 96.5 & 96.4 & 93.6 & 95.5\\
    SVRE-Admix-TI-DIM & \textbf{98.6} & \textbf{98.3} & \textbf{96.8} & \textbf{97.9}\\ 

\hline
\end{tabular}}
\end{center}
\end{table}

\vfill
~~
\vspace{20 em}
\vfill
~~
\vfill
~~
\vfill
~~
\vfill
~~


\end{document}